# A Novel Hybrid Deep Learning Approach for Non-Intrusive Load Monitoring of Residential Appliance Based on Long Short Term Memory and Convolutional Neural Networks

Sobhan Naderian[1]

[1] University of Glasgow sobhan.naderian@glasgow.ac.uk

**Abstract**

Energy disaggregation or nonintrusive load monitoring (NILM), is a single-input blind source discrimination problem, aims to interpret the mains user electricity consumption into appliance level measurement. This article presents a new approach for power disaggregation by using a deep recurrent long short term memory (LSTM) network combined with convolutional neural networks (CNN). Deep neural networks have been shown to be a significant way for these types of problems because of their complexity and huge number of trainable paramters. Hybrid method that proposed in the article could significantly increase the overall accuracy of NILM because it benefits from both network advantages. The proposed method used sequence-to-sequence learning, where the input is a window of the mains and the output is a window of the target appliance. The proposed deep neural network approach has been applied to real-world household energy dataset "REFIT". The REFIT electrical load measurements dataset described in this paper includes whole house aggregate loads and nine individual appliance measurements at 8-second intervals per house, collected continuously over a period of two years from 20 houses around the UK. The proposed method achieve significant performance, improving accuracy and F1-score measures by 95.93% and 80.93% ,respectively which demonstrates the effectiveness and superiority of the proposed approach for home energy monitoring. Comparison of proposed method and other recently published method has been presented and discussed based on accuracy, number of considered appliances and size of the deep neural network trainable parameters. The proposed method shows remarkable performance compare to other previous methods.

## 1. Introduction

Non-intrusive load monitoring is a well-structured approach that disaggregates the total electrical energy consumption of a house into its appliance level loads without the need of considerable metering devices on any appliance. Such issues could be challenging from the viewpoint of software-based detection for control of end-use patterns and growing internet of things (IoT) device. Recently, such challenge have been getting broad attention due to the potential advantages of energy disaggregation for the purposes of energy efficiency and the development of smart grid systems [1–3]

First researches on this topic started in 1985 by George Hart from the Massachusetts Institute of Technology (MIT) coined the term Non-intrusive (Appliance) Load Monitoring (NIALM) [4]. Basically, NILM implements combination of signal processing and machine-learning method to forecast the aggregate and individual appliance electricity consumption from electric power measurements in different locations of electric distribution network in residential regions).

Recently, researchers have been concentrated on NILM because of the availability of high technology and affordable measurement technologies (e.g., smart-grids) and also because of energy efficiency issues generated by the need to decrease the carbon pollutants of urban regions [5]. Besides, advances in machine learning methods and affordable processing cost, NILM technology is expected to serve as the backbone technology that will enable the creation of innovative smart-grid services [6]. The combination of the need for technologies promoting low-carbon emissions and the advances in machine learning and statistical techniques is presented in [7].

Several researches have been published about NILM analysis methods in recent years. Combination of the subtractive clustering and the maximum likelihood classifier presented in in [8], although, this approach only concentrated on type I load. The feasibility of using a temporal multi-label classification approach investigated by [9], statistical measures like mean value and variance are applied to identify each individual appliance footprint. The work reported on [10] proposed new method that combines wavelet transform and machine learning to nonintrusive load monitoring. Decision tree and wavelet coefficients of length-6 filter have been used for the proposed approach. Current waveform characteristic extraction using S-tranform and result identification by 0-1 knapsack method have been presented by [11]. An event-driven convolutional neural architecture for NILM of residential appliance has been introduced in [12]. The event-driven method is carried out to find the start-stop time and state change of the appliance precisely, including zero cross detection, current similarity detection, threshold evaluation, and event current acquisition.

Application of Karhunen-Loéve expansion method for splitting the active power signal into subspace components have been discussed in [13]. An appliances matching algorithm was presented to identify the turned-on appliance and an energy estimation algorithm was introduced for energy disaggregation. A new hybrid approach based on fusion of multiple time-domain features for appliance fingerprints extraction and a dimensionality reduction scheme has been presented for NILM by [14]. Authors used effective decision tree classifier for precise classification of electrical appliances by reduced features.

Based on published researches, the frameworks have been applied for the NILM problem are implemented using different combination of data preprocessing, feature extraction, and load recognition. Features extracted from the big electricity consumption data, then machine learning models like artificial neural networks (ANN) or pattern recognition algorithms utilized for individual appliance type detection [15]–[18]. NILM using convolutional neural networks (CNN) with differential input has been carried out by authors of [19]. The differential input extracts the transient feature hierarchically through five convolutional layers and then concatenated with the auxiliary input. In [20] the application of the pinball quantile loss function for deep neural network guidance has been discussed to carry out NILM. The proposed architecture of this paper enhances neural networks like Convolution Neural Networks and Recurrent Neural Networks.

Based on mentioned articles, accurate detection type of appliances, connected to the smart grid is very significant because it provides end-users with precise appliance grade utilization footprint. Accurate appliance behavior identification make better plan for demand side management and consumption pattern adjustment. To take account of these goals, this paper proposes a new method for supervised electric power consumption disaggregation.

New method proposes combined architecture of a deep recurrent neural network (RNN) based on the long short term memory (LSTM) architecture and convolutional neural network (CNN). The task of the network is to extract the target power signal of the appliance from the aggregate power signal based on LSTM and CNN after a supervised training of the network.

The paper is organized as follows. In section 2, the architecture of the LSTM and CNN is briefly described. Section 3 explains how to use the network for supervised non-intrusive load monitoring. In section 4 detailed output results of the network for the REFIT dataset and comparison with other proposed approaches have been discussed.

## 2. Deep Learning Neural Networks

### 2.1 Recurrent Neural Network (RNN) Networks

Recurrent neural network (RNN) is a class of artificial neural networks where connections between nodes form a directed graph along a temporal sequence. This allows it to exhibit temporal dynamic behavior. Derived from feed forward neural networks, RNNs can use their internal state (memory) to process variable length sequences of inputs. This makes them applicable to tasks such as time series, handwriting recognition and speech recognition.

A RNN with feedback in time is suitable to learn any dynamic (time-varying) mapping from input to output [21, 22]. Hence RNN suits better for power disaggregation than NN because in NILM applications, the power signal of almost all appliances is not deterministic due to random switch-on/off and the aggregate power signal is dynamic due to a time-varying superposition of power signals of different appliances.

#### 2.1.1 Long Short Term Memory (LSTM) Network architecture

RNN's have troubles about the short-term memory. Too long sequences make issues for RNN in process of carrying information from earlier time steps to later ones. So if RNN trying to process a long time series for predictions, it may miss important information from the beginning. Hence, mentioned disadvantages of RNN imply the need of Long Short Term Memory (LSTM) which is a special kind of RNN's, capable of learning long-term time series. Unlike standard feed forward neural networks, LSTM has feedback connections. It can, not only process single data points, but also entire sequences of data. LSTM's are capable to remember the information for a long period of time.

The LSTM model is shown in Figure 1 which is a sequential model that carries out by sequencing time series input data vector and provides output vector. The time-series data sent to a cell in sequential vector, at each step the cell output value is concatenated with next time step data and the output value of cell work as input for the next time step. This process is repeated until the last time step data.

LSTM was propsed by F. Gers [23] comprising of interactive neural networks, each of them consisting forget gate, input gate, input candidate gate, and output gate as shown in Figure 1. The output vector of the forget gate deviates between zero and one. The forget gate function which forgets the cell state from a previous time step that is not required and keep the necessary information cell state for prediction showed as

$$f_t = \sigma(W_f \cdot [h_{t-1}, x_t] + b_f) \quad (1)$$

the $\sigma$ function shows activation function usually called sigmoid which enables nonlinear characteristic of the model:

$$\sigma(X) = \frac{1}{1 - e^{-x}} \quad (2)$$

Afterwards, the input gate and input candidate gate activate together to make a new cell state $C_t$ which shifts to the next time step to update cell state. Sigmoid activation function and hyperbolic tangent function are applied as activation function at input gates and input candidate gate accordingly to generate new output and new cell state $C'_t$ described by the equations:

$$i_t = \sigma(W_i \cdot [h_{t-1}, x_t] + b_t) \quad (3)$$

$$C'_t = \tanh(W_c \cdot [h_{t-1}, x_t] + b_c) \quad (4)$$

The $tanh$ function is a hyperbolic tangent function that maps between -1 and 1.

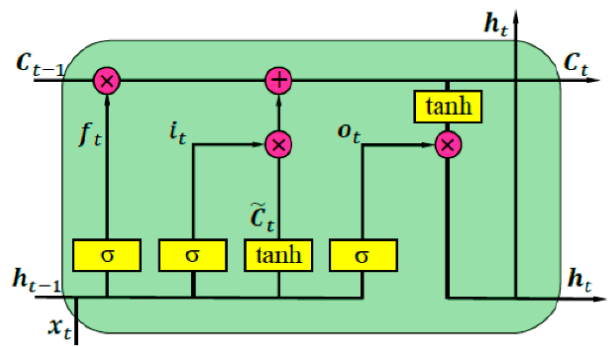

Figure 1. Internal structure of the LSTM cell

## 2.2. CNN

In deep learning, a convolutional neural network (CNN, or ConvNet) is a class of deep neural networks, most commonly applied to analyzing visual imagery. They are also known as shift invariant or space invariant artificial neural networks (SIANN), based on the shared-weight architecture of the convolution kernels that scan the hidden layers and translation invariance characteristics. Convolutional neural networks have been powerfully applied in different research topic and many scientific fields. CNN is consisted of numerous filters or kernels that working as features identifiers. Filters can detect specific short sequence and therefore be used to detect the specific variation sequence coming from the target appliance and filter out that coming from the non-target appliance. The parameters of the filters are automatically learned by CNN during the training [19].

### 2.2.1 CNN Architecture

A convolutional neural network consists of an input layer, hidden layers and an output layer. In any feed-forward neural network, any middle layers are called hidden because their inputs and outputs are masked by the activation function and final convolution. In a convolutional neural network, the hidden layers include layers that perform convolutions. Typically this includes a layer that applies multiplication or other dot product, and its activation function is commonly ReLU. Convolutional neural network layer types mainly include three types, namely convolutional layer, pooling layer and fully-connected layer. Figure 2 shows the CNN architecture:

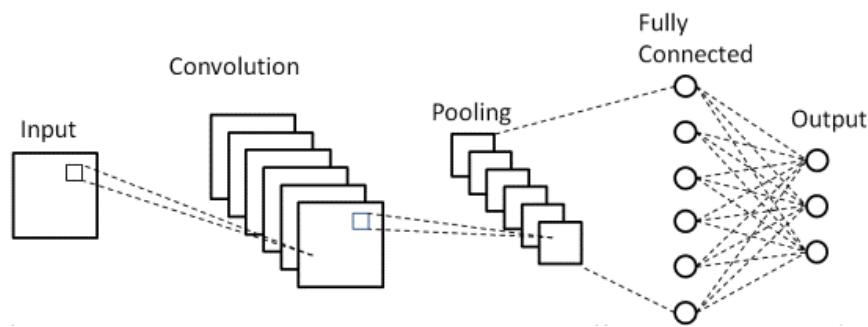

Figure 2. The architecture of CNN

### 2.2.2. Convolutional Layer

Convolutional layer is the crucial part of the CNN, which has local connections and weights with input features. The aim of Convolutional layer is to learn feature representations of the inputs. As shown in figure 2, Convolutional layer includes several feature maps. Each neuron of the same feature map is used to extract local features of different parts in the prvious layer, but for single neurons, its extraction is local characteristics of same positions in former different feature map. As a means to acquire a new feature, the input feature maps are first convolved with a learned kernel and then the results are sent into a nonlinear activation function. Different feature maps will be generated by applying different kernels. The typical kernel functions are sigmoid, tanh and Relu [25].

### 2.2.3. Pooling Layer

The pooling layer plays significant role to extract features from input data; it can decrease the dimensions of the feature maps and enhance the efficiency of feature extraction. It is usually placed between two Convolutional layers. The size of feature maps in pooling layer is calculated based on the moving step of kernels. The typical pooling operations are average pooling[26] and max pooling[27]. It is possible to extract more complex characteristics of inputs by cascading several Convolutional layer and pooling layer.

### 2.2.4. Fully-connected Layer

Generally, the classifier of Convolutional neural network could be one or more fully-connected layers. It takes all neurons in the previous layer and connects them to every single neuron of current layer. The last fully-connected layer is followed by an output layer. For classification purposes, softmax regression is commonly used because of proper probability distribution generating for the outputs. Support Vector Machine (SVM) is another well-known approach, which can be combined with CNNs for classification purposes [28].

## 3. The Proposed Hybrid Deep Learning Framework

In this section, the parameters considered for the proposed model will be discussed. Then, the architecture of the proposed deep learning model will be presented. The convolutional layers are used to extract the valuable features from the input data and then LSTM layers are used to forecast specified appliance load data sequence accurately.

### 3.1 Preprocessing and Preparing Data

In this part sequence to sequence model is applied for learning and testing of the proposed model. The preprocessing approach creates a window that slides on the input data (main utility data points) and try to forecast the output appliance load data points properly. The input window size is same for each appliances and the proposed deep learning model forecast same output window for each step. So the preprocessing algorithm generates matrices for input and output data points. Then the normalization is applied for both input and output data the preprocessing method subtract average of data from each data point and divide them by standard deviation as follows:

$$DP_{i_{Norm}} = \frac{DP_i - DP_{avg}}{DP_{std\_dev}} \quad (5)$$

where $DP_{i_{Norm}}$ stand for normalized data point, $DP_i$ is raw data point, $DP_{avg}$ and $DP_{std\_dev}$ are used for calculating the total average and standard deviation of data points, respectively.

## 3.2 Proposed Model Architecture

The architecture of the proposed deep learning framework has been presented in Figure 3. The CNN feature extraction block consists of a 1D convolutional layer and a MaxPooling layer. The convolutional approach is very powerful because it enables the first layer to learn about hidden features in the input data points. . A pooling layer is added after the convolutional layer for alleviating the limitation of the invariance of the produced feature map. In proposed model, a MaxPooling layer has been used which is a down-sampling approach that decreases the spatial dimension of the feature maps by a factor of 2, so minimize the total computational burden.
Two LSTM layers are connected to the CNN layer for forecasting the next sequence of output data accurately. Finally two dense layers added to convert the output of LSTM layers into proper output data.

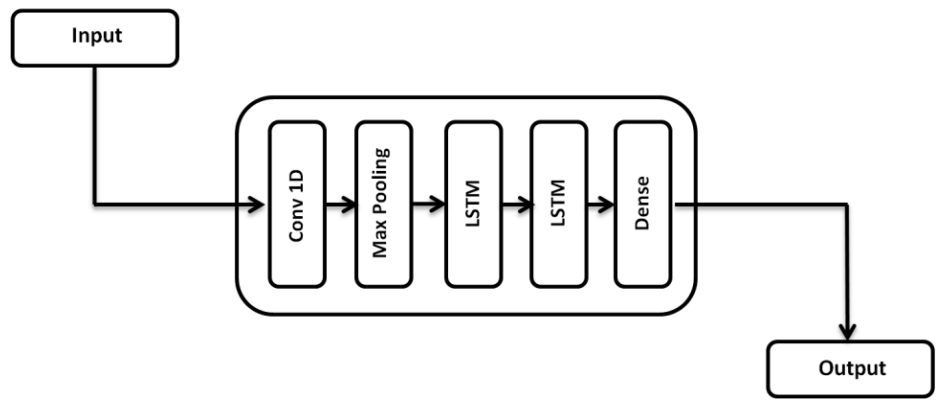

Figure 3. The architecture of proposed deep neural network

## 4. Results and Discussion

In this section, the output results of the proposed NILM algorithm based on hybrid deep learning neural network architecture, have been presented. Different evaluation stages are performed in order to validate the proposed NILM system. In this section, each part is outlined separately. First and foremost, the properties of power consumption databases have been described.

### 4.1 Dataset

The REFIT dataset [29], has been used in this article to test and validate proposed model. The REFIT dataset is one of the largest UK datasets that contains active power measurements, sampled at 8 sec resolution, and collected over a continuous period of 2 years from 20 UK homes. Five deep neural networks are trained to disaggregate five types of appliances as follows: fridge (FR), dishwasher (DW), microwave (MW), washing machine (WM) and kettle (KT) in house 2 of the REFIT dataset.

### 4.2 Proposed Hybrid Deep Learning Parameters

The architecture of the proposed deep learning network has been discussed in section 3.2. Parameters of proposed network have been selected using different types of architecture for better accuracy. Also part of architecture has been presented in [30] has been considered as well. The configuration of all layers of proposed model is depicted in Table 1. 100,000 data points have been used for each appliance during train and test procedure. 70 % of data points are used for training the proposed method and rest of them applied for the test of it. The proposed method is trained 500 times to predict appliance power consumption accurately. Sliding window with the size of 100 data points has been implemented for the input layer of date from main utility consumption data points. In this model, kernel size of 48 is considered for the first CNN-1D layer. Then *Max pooling* layer with pool size of 3 has been considered. Two LSTM layers with size of 128 and 256 are connected to second layers. The return sequence is set to true for these layers thus the network will output the full sequence of hidden states. Finally, two fully connected layers with size of 128 and window size are connected, respectively to generate the output data points. Mean absolute error (MAE) has been used as a loss function to monitor the loss.

Table 1. Parameters of the proposed deep neural network

| Proposed Model | |
|---|---|
| Convolution Layer | Kernels: 48 |
| | Input Shape: [100 , 1] |
| MaxPooling | pool size: 3 |
| LSTM1 | Hidden Nodes: 256 |
| | Return Sequence: True |
| LSTM2 | Hidden Nodes: 128 |
| | Return Sequence: False |
| Fully Connected Layer | Hidden Nodes: 128 |
| Fully Connected Layer | Hidden Nodes: 100 |

## 4.3 Performance Evaluation

Four metrics have been used to evaluate the accuracy and efficiency of the proposed method. Root Mean Square Error (RMSE) and average normalized error (ANE) have been calculated for each appliance as follows:

$$RMSE = \sqrt{\frac{1}{N}\sum_{t=1}^{N}(x_t - \hat{x}_t)} \qquad (6)$$

where $x_t$ is the real value, $\hat{x}_t$ is the prediction of an appliance at time $t$ and $N$ stands for number of non-missing data points. To measure average normalized error (ANE) this expression should be considered:

$$ANE = \frac{\left|\sum_{i=1}^{N}\bar{p}_i - \sum_{i=1}^{N}\hat{p}_i\right|}{\sum_{i=1}^{N}\bar{p}_i} \qquad (7)$$

This metric calculates the total energy difference between power forecasted by the NILM algorithm and the actual power utilized, across all 5 types of appliances. Where $\hat{p}_i$ is the power forecasted by the proposed algorithm from all 5 types of appliances at time $i$ and $\bar{p}_i$ is the actual power consumed by all mentioned appliances at the same time.

The accuracy and F1 score metrics have been implemented in the proposed framework to assess the performance of the method. They are measured as follows [31]:

$$Accuracy = \frac{TP + TN}{TP + TN + FP + FN} \qquad (8)$$

$$F1\ score = 2 \times \frac{precision \times recal}{precision + recal} \qquad (9)$$

where $precision = \frac{TP}{TP+FP}$ and $recal = \frac{TP}{TP+FN}$. $T, TN, FP$ and $FN$ depict the number of true positives, true negatives, false positives and false negatives, respectively. The number of time slices in which appliance $n$ was either correctly classified as being on (TP), classified as being on while it was actually off (FP), classified as on while it was actually on (FN) and correctly classified as being on (TN).

$$TP^{(n)} = \sum_t AND\ (x_t^{(n)} = on,\ \hat{x}_t^{(n)} = on)  \quad (10)$$

$$FP^{(n)} = \sum_t AND\ (x_t^{(n)} = off,\ \hat{x}_t^{(n)} = on)  \quad (11)$$

$$TN^{(n)} = \sum_t AND\ \left(x_t^{(n)} = off,\ \hat{x}_t^{(n)} = off\right)  \quad (12)$$

$$FN^{(n)} = \sum_t AND\ \left(x_t^{(n)} = on,\ \hat{x}_t^{(n)} = off\right)  \quad (13)$$

### 4.4 Results and Discussion

As discussed before, five deep learning neural networks are trained to disaggregate fridge (FR), dishwasher (DW), microwave (MW), washing machine (WM) and kettle (KT). Each network is trained to forecast the specified appliance output power consumption for house 2 in REFIT dataset.

Table 2 shows the accuracy and efficiency metrics for each appliance in detail. The proposed method shows the considerable accuracy for identification of appliances' power consumption, particularly for microwave and dishwasher. Figure 4 represent the output of proposed method for each appliance as well as real value of power consumption for each of them. Fridge power consumption is identified accurately because it works 24 hours while other appliances could have been used randomly and without specified pattern. Washing machine has heating/washing and washing processes. The proposed method has identified heating/washing process accurately. Because of complex pattern of washing part, it is identified with error but overall accuracy for whole process is significant. Usually, microwave and kettle are used for a short period of time in houses but proposed method identified them remarkably.

Table 2 Output result for each appliance

| Metrics | Fridge | Washing Machine | Dish Washer | Microwave | Kettle | Overall |
|---|---|---|---|---|---|---|
| ANE | 0.0374 | 0.0207 | 0.0303 | 0.7176 | 0.2110 | **0.2034** |
| RSME | 19.0667 | 297.8966 | 201.9924 | 97.8498 | 666.7836 | **256.71782** |
| Accuracy | 94.2105 % | 95.1405 % | 97.7904 % | 99.1509 % | 93.3814 % | **95.93474 %** |
| F1 score | 90.1170 % | 96.1575 % | 94.5996 % | 47.6190 % | 76.1904 % | **80.9367 %** |

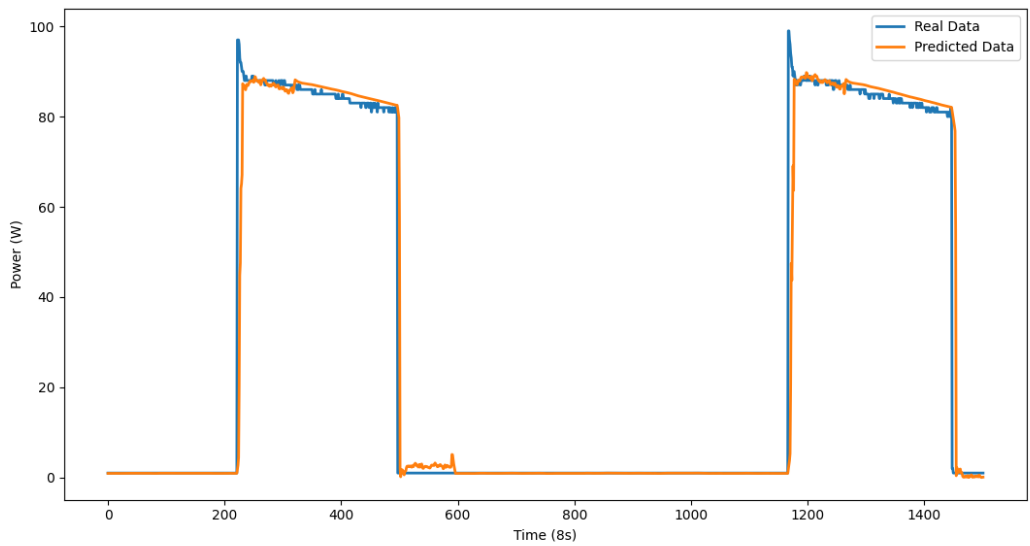

**Figure 4. Output results for fridge**

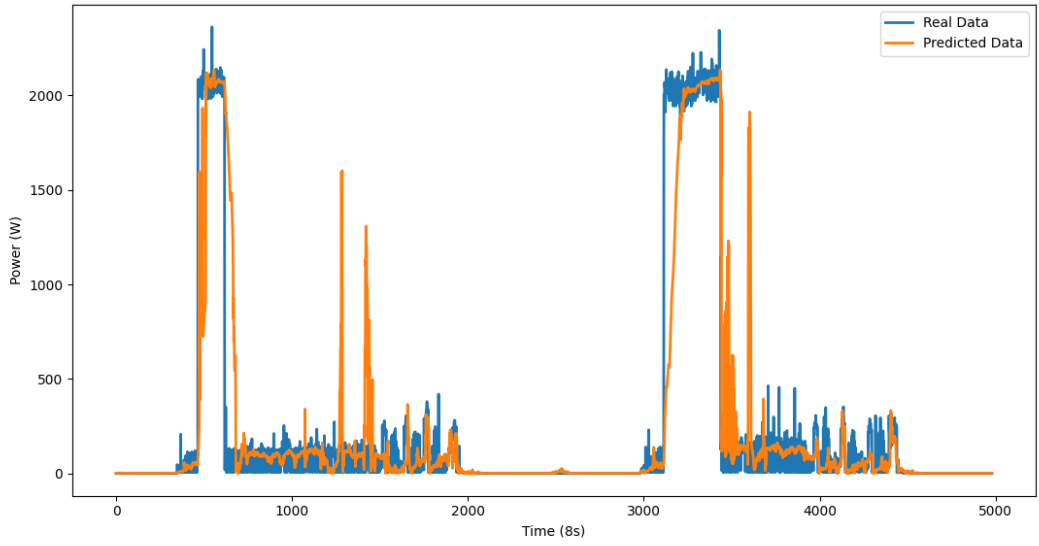

**Figure 5. Output results for washing machine**

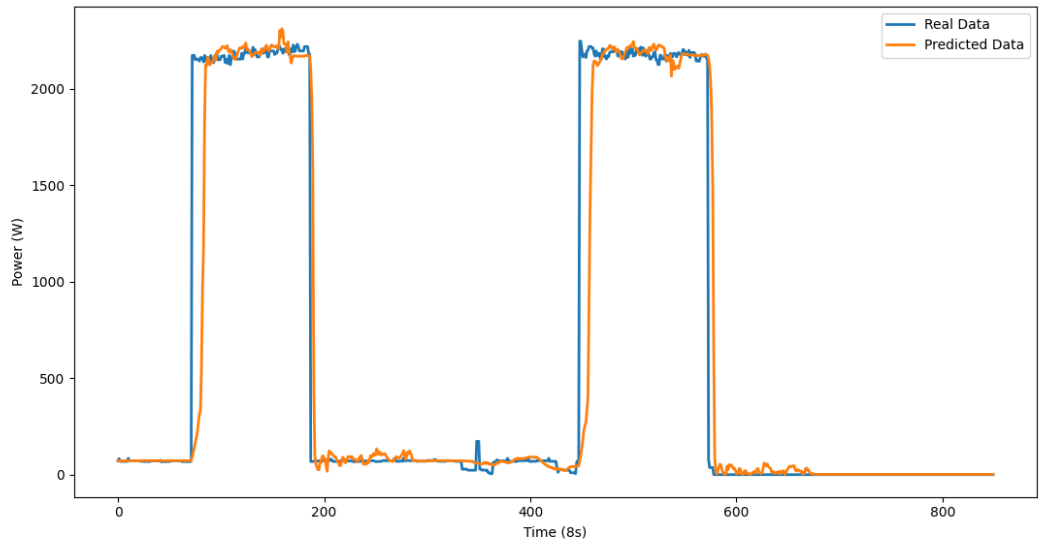

**Figure 6. Output results for dish washer**

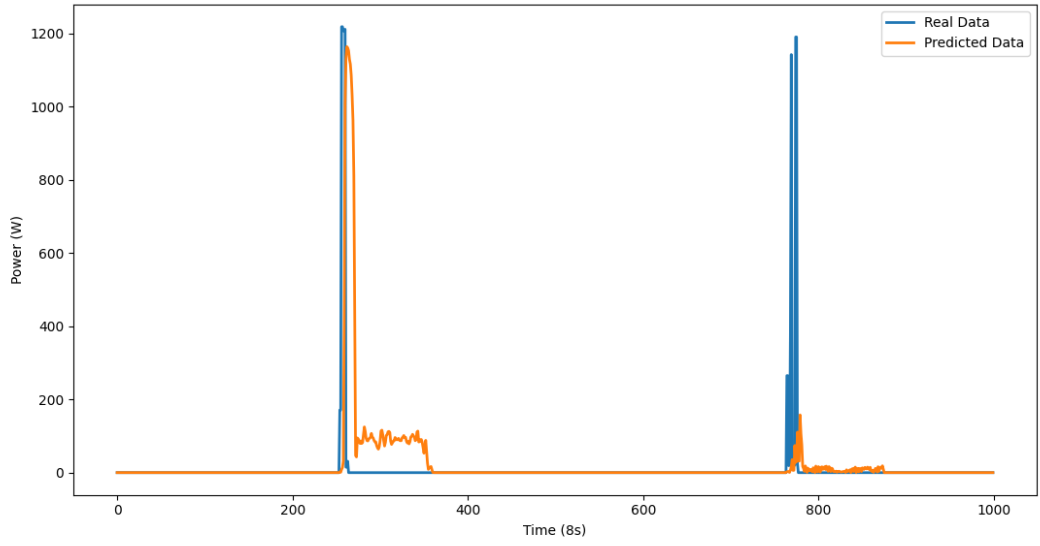

**Figure 7. Output results for microwave**

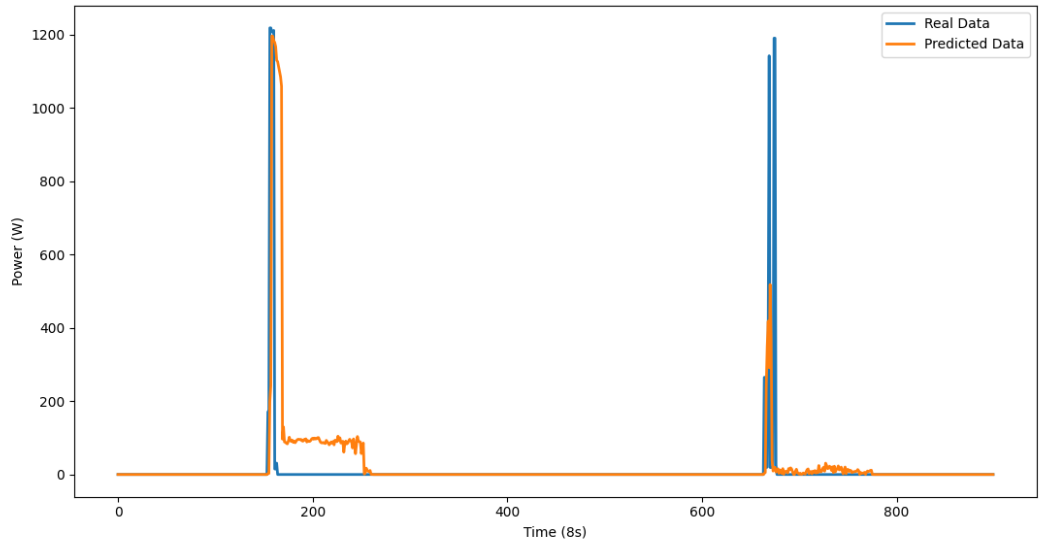

**Figure 8. Output results for kettle**

## 4.5 Discussion

The application of a public dataset allows assessing the outcomes obtained in comparison with other methods published recently. In order to compare the results of the proposed method for NILM algorithm based on hybrid combination of CNN and LSTM neural networks with the previous methods, the accuracy criteria is selected for comparison, as presented in Table 3. The previous NILM algorithms are taken from [31], [32], and [33] to compare the accuracy of them with proposed method.

These articles are evaluated based on different items, including (i) their implemented method, (ii) dataset that used, (iii) the number of appliance used and (iv) the accuracy. The results in Table 3, show that the accuracy of the proposed algorithm is superior to the identification accuracy from reference [32] and [33], and slightly higher than [34]. The accuracy in [34] is slightly lower compare to proposed method, because it is trained for more appliances compare to the proposed method. [33] and proposed method used same dataset but the accuracy of the proposed method significantly higher, although [33] trained for 6 appliances but differences between accuracies are high.

Table 3. comparison of proposed method and selected published method

|  | Method | Dataset | # Appliances | Accuracy (%) |
|---|---|---|---|---|
| **[32]** | Hidden Markov model | REDD | 5 | 88.6 % |
| **[33]** | Graph Signal Processing | REFIT | 6 | 71 % |
| **[34]** | CNN | UK-DALE | **7** | 93.8 % |
| **Proposed Method** | LSTM-CNN | REFIT | 5 | **95.93 %** |

To compare proposed framework from different viewpoint, comparison between sizes of parameters for other deep learning models could be useful. Table 4 shows the comparison based on size of model for 3 other models. The size of proposed model (1.2M) is one or two orders of magnitude smaller than other deep learning models. In [35] sequence to sequence based CNN network has been used for disaggregation but the size of trainable parameters is too big so obviously it is a slow method and consumes lots of computation resources. Although [36] and [37] have smaller sizes of parameters compare to the [35] but the size of proposed method's parameters is the smallest and highly likely is the fastest.

Table 4. Comparison between sizes of the parameters

| Model | # Parameters |
|---|---|
| CNN [35] | 29.8 M |
| LSTM [36] | 2.7 M |
| CNN [37] | 3.2 M |
| Proposed Method | **1.2 M** |

## 5. CONCLUSIONS

This article investigates the feasibility of a hybrid combination of deep recurrent LSTM and CNN neural network for power disaggregation in NILM applications. Proposed method is capable to forecast the power signal of five different types of appliances from the aggregate signal after a supervised training of the network by using available input data of the specified appliance in REFIT dataset. The major advantages of the proposed method are its accuracy and the capability to disaggregate the characteristics of each appliance. The validity of the proposed method have been presented and compared with other methods. The topic of applying the proposed algorithm to real project is worth further studying and testing.